\pdfoutput=1

\documentclass[11pt]{article}

\usepackage[preprint]{acl}

\usepackage{amsmath,amsfonts,bm}

\def\eqref#1{equation~\ref{#1}}

\def\1{\bm{1}}

\DeclareMathAlphabet{\mathsfit}{\encodingdefault}{\sfdefault}{m}{sl}
\SetMathAlphabet{\mathsfit}{bold}{\encodingdefault}{\sfdefault}{bx}{n}

\usepackage{times}
\usepackage{latexsym}
\usepackage{adjustbox}
\usepackage[symbol]{footmisc}

\usepackage[T1]{fontenc}

\usepackage[utf8]{inputenc}

\usepackage{microtype}

\usepackage{inconsolata}

\usepackage{graphicx}

\newcommand\blfootnote[1]{%
  \begingroup
  \renewcommand\thefootnote{}\footnote{#1}%
  \addtocounter{footnote}{-1}%
  \endgroup
}

\title{EpMAN: Episodic Memory AttentioN for Context Length \\Extension in Language Models}
\title{EpMAN: Episodic Memory AttentioN for Generalizing to Longer Contexts}

\author{Subhajit Chaudhury\textsuperscript{*}, Payel Das\textsuperscript{*}, Sarathkrishna Swaminathan, Georgios Kollias, Elliot Nelson, \\{\bf Khushbu Pahwa, Tejaswini Pedapati, Igor Melnyk\textsuperscript{\textdagger}, Matthew Riemer}  
\\subhajit@ibm.com, daspa@us.ibm.com, sarath.swaminathan@ibm.com, \\gkollias@us.ibm.com, enelson@ibm.com, kp66@rice.edu,  tejaswinip@us.ibm.com, \\igor.melnyk@capitalone.com, mdriemer@us.ibm.com\\
\\IBM Research}

\begin{document}
\maketitle
\begin{abstract}
\blfootnote{\textsuperscript{*} denotes equal contribution; \textsuperscript{\textdagger} Work done at IBM Research}
Recent advances in Large Language Models (LLMs) have yielded impressive successes on many language tasks. However, efficient processing of long contexts using LLMs remains a significant challenge. We introduce \textbf{EpMAN} -- a method for processing long contexts in an  \textit{episodic memory} module while \textit{holistically attending to} semantically relevant context chunks. The output of \textit{episodic attention} is then used to reweigh the decoder's self-attention to the stored KV cache of the context during training and generation. When an LLM decoder is trained using \textbf{EpMAN}, its performance on multiple challenging single-hop long-context recall and question-answering benchmarks is found to be stronger and more robust across the range from 16k to 256k tokens than baseline decoders trained with self-attention, and popular retrieval-augmented generation frameworks.
\end{abstract}

\begin{figure}[tb]
    \centering
    \includegraphics[origin=c,width=0.45\textwidth]{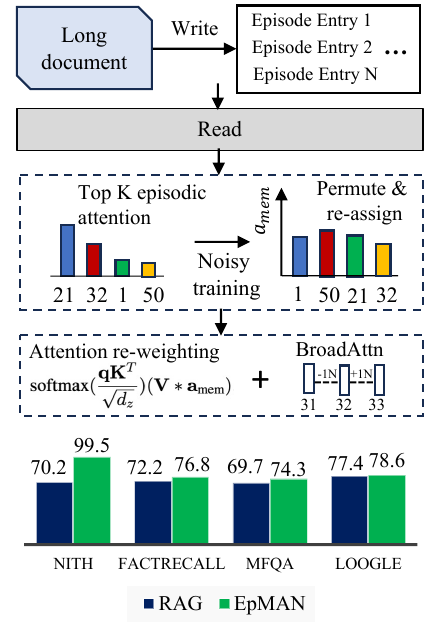}
    \caption{\textbf{EpMAN} uses episodic attention and noisy training for robust long context performance on recall and QA tasks (mean over 16k - 256k context lengths)}
    \vspace{-0.5cm}
    \label{fig:front_fig}
\end{figure}

\section{Introduction}
\label{sec:introduction}

Large language models (LLMs) are highly capable of many natural language processing (NLP) tasks; however, they still struggle with generalization to long inputs that are unseen during training. To enhance the generalization ability of LLMs on unseen long inputs, continual pretraining on longer sequences has been attempted, which requires significant computational investments~\cite{abdin2024phi}.  One main challenge of 
training with long context is 
the quadratic memory and time complexity of the current self-attention mechanism employed by most LLMs,  
making it computationally expensive and infeasible for processing long sequences. To circumvent this, existing solutions often resort to techniques like sliding window attention, dilated sliding window, and sparse attention~\cite{beltagy2020longformerlongdocumenttransformer, child2019generatinglongsequencessparse}. 
In parallel, scalable position embeddings-based approaches, such as position interpolation and length extrapolation, have been proposed which involve minimal finetuning~\cite{chen2023extendingcontextwindowlarge}.

Despite recent advances in long context processing abilities of LLMs,  recent long-context modeling benchmarks show that LLMs still underperform in terms of modeling the input context that has a length longer or even similar to those seen during training~\cite{kuratov2024babilong, hsieh2024rulerwhatsrealcontext}. A promising solution to the problem of long context processing is the use of retrieval-augmented-generation (RAG) frameworks. 
RAG  combines the strengths of retrieval models and generative LLMs to handle long contexts. In this framework, a retrieval model first identifies and retrieves the relevant context from a large corpus, which is then passed to the generative model for text generation. Despite its usefulness, RAG struggles to handle situations where there remains conflict between retrieved information and parametric memory, or the retrieved context contains irrelevant information, resulting in hallucination or ignoring the context while answer generation~\cite{xieadaptive}.

Thus, the current gap in long context modeling of LLMs calls for alternative and efficient mechanisms for long context handling. For this purpose, in this work, we propose a second attention mechanism, named as episodic memory attention (\textbf{EpMAN}), shown in Figure~\ref{fig:front_fig}, which is utilized to scale the self-attention according to the importance of the information present in the context. Inspired by the dual processing theory proposed in~\cite{kahneman2011thinking}, in which self-attention can be characterized by “System 1”, a mode of thought that is fast, instinctive but less accurate, the proposed \textbf{EpMAN} can mimic the slow and calculative thinking steps,   i.e., the “System 2” mode. \textbf{EpMAN} considers writing text chunks from the context in an episodic memory module, estimating their relative relevance with respect to a given query, and then utilizing this relevance to reweigh the self-attention. Experiments on challenging fact recall and single-hop question-answering from long context scenarios, which include the presence of distractions and confusions, as well as replaced keywords and rephrased sentences in the input context, show the benefit of the LLM trained with \textbf{EpMAN}, compared to the LLMs trained on long inputs using self-attention and RAG frameworks.

Our main contributions are:
\begin{itemize}
    \item A novel architecture combining episodic memory attention with self-attention during LLM training, which is inspired by the dual processing theory.
    \item An effective training method that introduces noise while estimating attention to the relevant chunks stored in the episodic memory. 
    \item An attention scope expansion method employed during inference which enables attending to the broader context in a more holistic manner.
    \item The proposed framework shows better generalization to recalling and answering from challenging long context which includes information that is confusing, irrelevant, or contains replaced keywords or rephrases when compared to LLMs trained on long context and RAG systems.
\end{itemize}

\section{Related work}
\label{sec:related_work}

Increasing context length in LLMs introduces several challenges that impact model performance. We elaborate on some of those problems and also other memory-augmented techniques. 

\subsection{Long Context Challenges}
\textbf{Recency Bias}: 

Recent studies\cite{recency_bias1, recency_bias2, recency_bias3, recency_bias4} have shown that LLMs tend to prioritize information found towards the end of a context while neglecting important details presented in the beginning and the middle parts of the context. 
This bias is believed to originate from the pre-training process, where the most informative tokens for prediction are typically the most recent ones. 

To address this issue, the authors in \cite{attention_sorting} propose \textit{attention sorting}, which

rearranges documents based on their attention weights and moves documents that receive higher attention during decoding towards the end of the context.

\textbf{Impact of Distractors}: 
Another significant challenge is the impact of distractors as highlighted in \cite{attention_sorting}, the accuracy of long-context language models generally decreases as the context length increases through the addition of distractor documents~\cite{distractor2, distractor_rag, distract_long_context}. This stresses that an overabundance of information, even if irrelevant, can hinder the model's ability to identify and utilize the most pertinent parts of the context effectively.

\textbf{Attention Dilution}
Long-context modeling in LLMs also suffers due to the phenomenon of \textit{attention dilution}, explored in \cite{attention_dilution1,attention_dilution2,attention_dilution3,attention_dilution4} which occurs due to the softmax normalization in the attention mechanism. Since attention weights must sum to 1, the presence of many irrelevant documents can result in each receiving a small but non-negligible amount of attention. This dilution of focus can overshadow the model's ability to concentrate on the most crucial information.

To address this, the research in \cite{attention_distillation} proposes a strategy to mitigate attention dilution in RAG-based systems by training the retriever with attention scores from a fine-tuned reader. 

However, if the reader is not fine-tuned well, the attention scores it provides might be unreliable, leading to suboptimal retriever training and ultimately impacting overall performance. Additionally, distilled attention mechanisms might inadvertently amplify existing biases present in the training or retrieved data. Differential Transformer~\cite{ye2024differential} also aims to reduce the noisy attention on irrelevant tokens by using noise cancellation by subtracting attention values using two softmax outputs.

\subsection{Memory-Augmented Retrieval}
\textit{Memory-augmented retrieval} involves storing past contexts and knowledge in a non-trainable memory bank, allowing the model to retrieve chunk-level information~\cite{memlong, memllm, mem-tree}. By storing information as key-value pairs and utilizing a retrieval mechanism, the model can access relevant past contexts. This approach has the potential to mitigate the limitations of fixed context windows and improve the model's ability to handle long-range dependencies. However, relying solely on single-layer representations for retrieval might not be robust enough and can be unstable.

Our proposed approach, \textbf{EpMAN}, resolves the challenges of recency bias, distractors, and other limitations by storing long contexts in a dedicated memory module and selectively attending to semantically relevant chunks. Rather than focusing on the most recent inputs, \textbf{EpMAN} retrieves relevant information from the entire stored context, effectively addressing the "lost in the middle" phenomenon, where relevant information in the middle of long contexts is often overlooked. Additionally, the proposed differentiating attention mechanism with the denoising objective reduces the impact of distractors, ensuring robust information processing.

Closer to \textbf{EpMAN}, \cite{wu2022memorizingtransformers} combines the attention to top-k nearest neighbor with self-attention by using a learnable gate; however, our approach is simpler, more intuitive, and more suitable for long context generalization. Another memory-augmented LLM, known as Larimar~\cite{das2024larimar}, attends to the readout from an episodic memory storing the context during decoding and 
performs gradient-free write to the memory for input context length generalization. However, Larimar only attends to a single top-1 readout and therefore is not suitable for handling tasks in which relevant information is diffused over the context. 

\section{EpMAN: Episodic Memory AttentioN}
\label{sec:epman}

In this section, we first describe the standard attention implementation in transformer-based language models~\cite{attention_is_all_you_need}. Subsequently, we outline our proposed differentiating attention over the KV cache using the episodic memory, referred as \textbf{EpMAN}. The \textbf{EpMAN} mechanism enables focusing on the relevant information required for correct recall or answering, which can be diffused over the long context in practice. 

\subsection{Preliminaries}
The standard attention mechanism in LLMs is used to assign relevance weights to the input sequences when generating the output sequence. The model learns to pay attention to different tokens of the input sequence for each token of the response, enabling it to generate more accurate and relevant outputs to the context. The attention mechanism is implemented using a variant of the scaled dot-product attention mechanism as described below.

Let us denote the input sequence as $\mathbf{X} = [\mathbf{x}_1, \mathbf{x}_2, \dots, \mathbf{x}_n]$, where $\mathbf{x}_i$ is the $i$-th input vector, and $n$ is the length of the input sequence. The attention mechanism computes a set of attention weights $\mathbf{a} = [a_1, a_2, \dots, a_n]$, which sums to 1 and is a distribution over the input sequence. These attention weights are used to compute a weighted sum of the input vectors, which is then used as input to the decoder for the next token.

In the standard attention, we compute the query vector $\mathbf{q}$ as a function of the current decoder hidden state $\mathbf{h}_t$, $\mathbf{q} = f(\mathbf{h}_t)$, where $f$ is a linear transformation that maps the decoder hidden state to the query vector. Next, we compute the keys $\mathbf{K} = [\mathbf{k}_1, \mathbf{k}_2, \dots, \mathbf{k}_n]$ and values $\mathbf{V} = [\mathbf{v}_1, \mathbf{v}_2, \dots, \mathbf{v}_n]$, where $\mathbf{k}_i$ and $\mathbf{v}_i$ are linear transformations of the input vectors $\mathbf{x}_i$ similar to query vector. Then, we compute the dot product of the query vector $\mathbf{q}$ and each key vector $\mathbf{k}_i$, followed by a softmax which is then multiplied by the value vectors to get the context vector at that token as $c_i = \text{softmax}(\frac{\mathbf{q}\mathbf{K}^T}{\sqrt{d_z}})\mathbf{V}$. The query, key, and value vectors are learned during training.

\begin{table*}[]
\centering
\begin{adjustbox}{width=0.95\textwidth}
\begin{tabular}{l|llll|llll}
\hline
                                       & \multicolumn{4}{c|}{Paul Graham} & \multicolumn{4}{c}{PG19}      \\ \hline
Method                                 & 16k    & 32k    & 64k   & 128k  & 16k   & 32k   & 64k   & 128k  \\ \hline
Mistral-7B-Instruct-v0.2               & 62.10  & 82.35  & 45.10 & 25.40 & 89.00 & 99.55 & 59.95 & 30.00 \\
Phi-3-small-128k-instruct              & 26.40  & 56.00  & 72.00 & 89.65 & 15.60 & 56.35 & 75.75 & 71.55 \\ \hline
Dragon + Mistral-7B-Instruct-v0.2      & 71.70  & 68.80  & 72.50 & 67.75 & 80.80 & 81.80 & 82.90 & 87.10 \\
Dragon + Phi-3-small-128k-instruct     & 66.25  & 48.25  & 47.75 & 43.80 & 64.80 & 58.10 & 63.90 & 66.85 \\
\hline
EpMAN (uniform train - Exact test) & 100.0  &	100.0 &	99.2 &	97.9 &	99.5 &	100.0 &	99.5 &	100.0 \\
EpMAN (uniform train - NarrowAttn test) & 100.0	& 100.0 &	100.0 &	98.2 &	99.5 &	100.0 &	99.5 &	100.0 \\
EpMAN (noisy train - Exact test) & 100.0  &	100.0 &	99.1 &	97.9 &	\textbf{99.6} &	100.0 &	\textbf{99.6} &	100.0 \\
EpMAN (noisy train - NarrowAttn test) & 100.0	& 100.0 &	100.0 &	\textbf{98.3} &	\textbf{99.6} &	100.0 &	99.5 &	100.0 \\
\hline
\end{tabular}
\end{adjustbox}
\caption{Performance of various models on needle-in-the-haystack recall task with background / 
``haystack'' text from both sources - Paul Graham Essays and PG19. }
\label{NITH}
\end{table*}

\subsection{Details of EpMan - An Episodic Differentiating Attention}
\label{sec:epman_details}
While the standard attention mechanism in LLMs is effective for shorter contexts, it faces limitations when dealing with long context inputs due to issues like emergence of attention sink~\cite{xiao2023efficient}, conflict between input context and pretraining knowledge~\cite{xieadaptive,lveval}, and susceptibility to irrelevant information in context~\cite{borgeaud2022improving}. 
To address such challenges, we propose \textbf{EpMAN}, an episodic memory-based attention mechanism that enables finding relevant parts from the input context while discarding the irrelevant information, and then reweighing the standard attention to the relevant parts by using a relevance estimate. 

\textbf{Memory operations}: Given a large document as input, \textbf{EpMAN} first divides it into smaller entries (or chunks) that are written in the episodic memory. The memory consists of two operations, read and write. One can simply store encodings from a pretrained frozen retriever in the episodic memory as the \textit{write} operation, or train an MLP using the encodings as input for a learnable \textit{write} operation. Similarly, a learnable or a fixed (e.g., cosine) similarity function between the query encoding and the chunk encodings can be used to \textit{read} from the context. (more details on trainable read and write in the appendix)

We use cosine similarity for \textit{reading} and a state-of-the-art
pretrained retriever model named Dragon~\cite{lin2023train} in this work\footnote{We use the multi-turn version: nvidia/dragon-multiturn-context-encoder}. 
We refer to the score obtained from the read operation as episodic attention ($\textbf{a}_{mem}$) which is used to weigh KV cache attention for LM training.

\textbf{Replacing standard attention with differentiating attention}:
In addition to the latent retrieval encodings, the episodic memory also stores the KV cache of the context divided into episodic entries (stored in CPU memory due to increased size), which is processed using the above $\mathbf{a}_{\text{mem}}$ as follows. Once we get $\mathbf{a}_{\text{mem}}$ for each entry, we multiply the attention $\mathbf{a}_i = \text{softmax}(\frac{\mathbf{q}\mathbf{K}^T}{\sqrt{d_z}})$ with the $\mathbf{a}_{\text{mem}}$ episodic attention. This reweighing of standard attention with episodic attention $\mathbf{a}_{\text{mem}}$ provides the differentiating attention mechanism that focuses on the relevant chunk in the memory while discarding the irrelevant information in the context. It is important to note that $a_{\text{mem}}$ is the attention over chunks, so the attention is the same for all K-V token embeddings in the chunk. The resulting attention operation can be described as, 
\begin{equation}
    \textbf{a}_{epman} =  
\text{softmax}(\frac{\mathbf{q}\mathbf{K}^T}{\sqrt{d_z}})(\mathbf{V}*\mathbf{a}_{\text{mem}}),
\end{equation}
\noindent where we broadcast the $\mathbf{a}_{\text{mem}}$ value for each entry to the size of the number of tokens before multiplying with the value vector.

\textbf{EpMAN} thus provides a computationally efficient way to holistically handle long contexts in LLMs by leveraging an episodic memory attention mechanism. This approach allows the decoder model to attend to different chunks of the input sequence with different relevance estimates, which is used to self-distill the standard attention. This self-distillation of standard attention to input context enables generating more accurate and contextually relevant outputs.

\begin{figure}[tb]
    \centering
    \includegraphics[origin=c,width=0.48\textwidth]{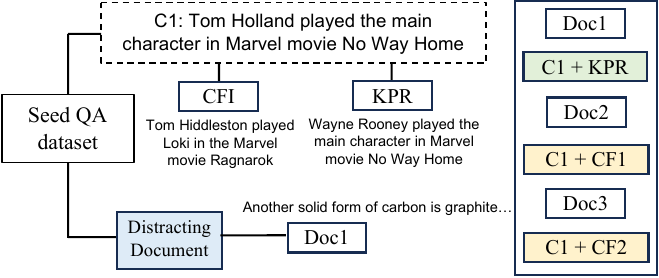}
    \caption{CFI and KPR in LV-Eval dataset.}
    \vspace{-0.2cm}
    \label{fig:lveval}
\end{figure}

\subsection{Synthetic Data for Training} 
To couple our decoder such that it follows the ranked $a_{mem}$ output from the memory operations, we train it on synthetic data. We use two kinds of training data as explained below:

 \subsubsection{Pre-training dataset} We train the model using a combination of the next token prediction task and memory retrieval task, following the loss objective in previous memory-enhanced architectures~\cite{das2024larimar}. We used synthetically generated passages from a teacher model (\texttt{mistralai/Mixtral-8x22B-Instruct-v0.1}) which serve as the context for the model. We then add distractor passages from Wikipedia in the context to increase the context length during training. We did not use hard negative mining for this data that we used for QA data described next.

 \subsubsection{QA synthetic data} We use two kinds of QA synthetic data as below,

  \textbf{Topic-sampled data:} To generate this dataset, we used the teacher model, providing it with a topic sampled from a predefined list. The model was tasked with generating a short paragraph based on the given topic, which could either be factual or fictional. Afterward, the same model was instructed to create two questions related to the passage: one that could be answered using the information from the text (a related question) and one that could not be answered solely using the text (an unrelated question). Additionally, the model generated answers for both questions. Finally, a verification step was performed using Llama-3-70B-Instruct model as a judge, along with the \texttt{nightdessert/WeCheck} consistency checker to ensure consistency between the generated passage, related/unrelated questions, and corresponding answers. 

  \textbf{Wikipedia}: Firstly, we randomly sample Wikipedia passages and generate questions and answers from these passages using a teacher model. We use few-shot examples to guide the teacher model in generating question answers.
 
 Similar to pretraining data, we add distractors from other Wikipedia passages in both cases above. In addition, to make the training more challenging, we mine context chunks that are similar to the topic of the relevant chunk (hard negatives) from a pool of Wikipedia entries which is added as part of the distracting context. We use an episode size of 16, where one of the entries is relevant and the others are distractors. Our chunk size is 256 and the effective training context size is 4K tokens. We use the index of the relevant chunk for episodic loss and the answer tokens for the decoder loss.

\subsection{Training with Denoising}
\label{sec:noisy_training}

The read operation assigns the episodic attention $\textbf{a}_{mem}$ on each of the entries and we use a threshold to keep the top K entries (similar to RAG) for that are seen by the decoder to answer the query. However, different from RAG, our method \textbf{EpMAN} allows for each of these entries to be weighted differently such that the decoder can learn from differentiating attention.

\begin{table*}[]
\centering
\begin{adjustbox}{width=0.85\textwidth}
\begin{tabular}{l|ccccc|c}
\hline
Method                             & 16k  & 32k  & 64k  & 128k & 256k & Mean \\ \hline
Mistral-7b-instruct-v0.2           & 65.3 & 72.5 & 41.0 & 22.5 & 11.5 & 42.6 \\
Phi-3-small-128k-instruct          & \textbf{82.0} & \textbf{80.5} & \textbf{81.0} & 63.0 & 34.5 & 68.2 \\ 
Dragon + Mistral-7b-instruct-v0.2  & 74.2 & 71.8 & 66.0 & \textbf{77.2} & 69.0 & 71.7 \\
Dragon + Phi-3-small-128k-instruct & 71.8 & 70.5 & 68.0 & 76.0 & 68.5 & 71.0 \\ 
\hline
EpMAN (Uniform train - Exact test)                        & 44.5 & 49.0 & 48.0 & 50.2 & 43.5 & 47.0  \\
EpMAN (Uniform train - NarrowAttn test)                        & 59.5 & 64.5 & 62.5 & 69.0 & 59.5 & 63.0 \\
EpMAN (Uniform train - BroadAttn test)                       & 82.0 & 73.0 & 71.5 & 70.0 & 79.0 & 75.1 \\
EpMAN (Noisy train - Exact test)                        & 44.5 & 49.0 & 51.0 & 51.2 & 45.5 & 48.2  \\
EpMAN (Noisy train - NarrowAttn test)                        & 60.2 & 64.5 & 61.8 & 68.5 & 59.0 & 62.8 \\
EpMAN (Noisy train - BroadAttn test)                       & 81.8 & 75.2 & 76.0 & 75.2 & \textbf{80.2} & \textbf{77.7} \\ \hline
\end{tabular}
\end{adjustbox}
\caption{Performance of various models on Factrecall-en using recall metric. \textbf{EpMAN} with noisy training with BroadAttn shows the overall best performance.}
\label{FACTRECALL}
\end{table*}

\textbf{Out-of-domain mismatch}: A straightforward method for decoder training would be to keep the original $\textbf{a}_{mem}$ weight that the read operation provides to each of the entries of the episode. However, this strategy is not always the best, 
particularly when the goal is to generalize to out-of-distribution samples. During training the decoder might become biased to expect the relevant chunk to always have a high episodic attention. For out-of-distribution (OOD) data, the read operation might not always assign the highest weight to the most relevant episodic entry, and in practice, the most relevant chunk might be ranked as one of the lowest in the top-K set of chunks. In that case, the above training strategy, would lead to poor generalization.

\textbf{Robust training with noisy attention}: Since the episodic us to assign different importance to each entry, 
\textbf{EpMAN} proposes a \textit{noisy} training scheme where the top K chunks receive random weights between 1.0 and 0.9. Throughout this work, we use K=5, unless otherwise mentioned. 
The episodic entries are further randomly permuted to change their relative order to ensure that they are not arranged in descending order of $a_{mem}$ weights. This training allows the decoder to pick up the relevant chunk even if it is not in the higher bins of the top K entries. This noisy training provides a denoising objective that allows the decoder to be robust compared to uniform $\textbf{a}_{mem}$ training.

\textbf{Loss}: We use two losses during training. The first loss is the episodic attention loss that minimizes the distance between the distribution of the episodic attention from the read operation and the true distribution of chunk relevance using cross-entropy loss for the case where the read and write operations are learnable. We also use the next-token prediction loss in the decoder. The total loss is  

\begin{equation}
    L = \mathbb{E}_\mathcal{D} [ \alpha \ln p(\textbf{l}|\textbf{q}, \textbf{C}) + \ln(\textbf{a}|\textbf{q}, \textbf{C}, \textbf{a}_{mem})],
\end{equation}
\vspace{1em}
where ${(\textbf{q}, \textbf{C}, \textbf{l}, \textbf{a}) \sim \mathcal{D}}$ represent the query, context, location of relevant episodic entry and decoder response respectively. We use $\alpha$ as the weight for the episodic loss which is typically set to 0.1. In the main paper, we report results with only decoder loss for fair comparison with RAG systems.

\subsection{BroadAttn: Neighborhood expansion during Inference}
The top-K episodic entries might be arranged in a manner such that there might be information cutoff during read operation (for e.g. delayed co-reference). In such cases, which is referred to as the \textit{NarrowAttn} as the decoder's attention only includes top-K chunks, the subject might be described in an entry (e.g. ``Albert Einstien was born in Germany'') whereas some attribute related to the subject might be described in a separate entry (e.g. ``He taught himself algebra''). To improve the robustness of \textbf{EpMAN} in such cases, we expand episodic attention during inference such that it includes the immediate (sequential) neighbors of each of the top-K chunks, which is referred as the \textit{BroadAttn}. We also test the setting where exact $\textbf{a}_{mem}$ weight for each chunk is attended during test, referred to as \textit{exact test}. 

Both \textit{NarrowAttn} and \textit{BroadAttn} consider preserving the original order of the chunks in which the information is presented in the original context, as suggested by~\cite{yu2024defense}. 

\begin{table*}[]
\centering
\begin{adjustbox}{width=0.85\textwidth}
\begin{tabular}{l|lllll|l}
\hline
{Method} &
  {16k} &
  {32k} &
  {64k} &
  {128k} &
  {256k} & 
  {Mean} \\ \hline
Mistral-7b-instruct-v0.2              & 65.3  & 52.5 & 37.6 & 34.7 & 25.0 & 43.0 \\ 
Phi-3-small-128k-instruct             & 45.5  & 52.5 & 49.5 & 31.7 & 33.7 & 42.6 \\ 

\hline
Dragon + Mistral-7b-instruct-v0.2      & {68.3}  & {71.3} & 66.3 & {71.3} & {71.3} & 69.7 \\ 
Dragon + Phi-3-small-128k-instruct     & 56.4  & 50.5 & 50.5 & 58.4 & 51.5 & 53.5 \\
\hline
EpMAN (Uniform train - Exact test)                        & 64.3  & 59.4  & 65.3  & 69.3  & 64.4  & 64.5  \\
EpMAN (Uniform train - NarrowAttn test)                        & 63.4 & 66.3 & 65.4 & 72.3 & 63.4 & 66.1 \\
EpMAN (Uniform train - BroadAttn test)                       & 69.3 & 70.3 & 71.3 & 68.3 & 72.3 & 70.3 \\
EpMAN (Noisy train - Exact test)                        & 59.4  & 60.4  & 62.4  & 64.4  & 60.4  & 61.4  \\
EpMAN (Noisy train - NarrowAttn test)                        & 71.3 & 66.3 & 67.3 & 75.3 & 63.4 & 68.7 \\
EpMAN (Noisy train - BroadAttn test)     & \textbf{74.3} & \textbf{73.3} & \textbf{73.3} & \textbf{75.3} & \textbf{75.3} & \textbf{74.3} \\ 
\hline
\end{tabular}
\end{adjustbox}
\caption{Performance of various models on multifieldqa-en-mixup using LLM-as-Judge.}
\label{MFQA}
\end{table*}

\section{Experimental details and results}
\subsection{EpMAN Implementation Details} We use {mistralai/Mistral-7B-Instruct-v0.2} as our decoder and Dragon~\cite{lin2023train} as the retriever for our experiments with \textbf{EpMAN}. We use a sequence length of 256 tokens for each entry and we cut at sentence boundaries. We use an effective batch size of 32 and train the models for 20k steps. We used LoRA~\cite{hu2021lora} for training. 

\subsection{Evaluation datasets}
\label{sec:eval_data}
We evaluate \textbf{EpMAN} on a combination of recall and question-answering tasks. For recall tasks, we evaluate on the Needle in the haystack (NITH)~\cite{needlehaystack_kamradt} and factrecall-en tasks. For question answering, we use two single-hop, single-document QA tasks: Multifield QA~\cite{bai2023longbench} and Loogle SD~\cite{li2023loogle}. 
For factrecall-en, Multifield QA, and Loogle SD, we use the \textbf{LV-Eval} benchmark, as proposed by \cite{lveval}, which subjects LLMs to a more challenging evaluation by inserting confusing facts into the context and by replacing keywords and phrases in the context to make sure that LLMs use comprehension of the context, rather than prior knowledge, to answer the questions. We consider the most challenging scenario included in the LV-Eval framework, where the context contains both confusing facts and replaced keyword and phrases. 

\textbf{Needle in the haystack}: We use NITH~\cite{needlehaystack_kamradt} which is a well-known recall task from long context inputs to LLM. This task assumes that there is a needle sentence located in the large context input (haystack) and it tests if an LLM can complete a partial representation of that sentence. We use context lengths of varying size (16k, 32k, 64k, 128k) with needles located in 200 evenly spaced locations. Our needle sentence is: "\emph{The best thing to do in San Francisco is } eat a sandwich and sit in Dolores Park on a sunny day" (with the partial representation in italics). We experiment with two haystacks: (i) The dataset of \textbf{Paul Graham} essays following \cite{needlehaystack_kamradt} and (ii) books from \textbf{PG19} corpora \cite{Rae2020Compressive}, inspired by \cite{kuratov2024babilong}. We concatenate the full set of Paul Graham essays and shuffle the sentences from all PG19 test texts\footnote{\url{https://huggingface.co/datasets/emozilla/pg19-test}.} (>11M tokens), prior to context selection.

\textbf{LV-Eval:} LV-Eval is a long context benchmark with the context length varying between 16k, 32k, 64k, 128k and 256k. LV-Eval is comprised of various datasets, including the Factrecall-en, Multifield QA and Loogle SD datasets. 
To begin with, it is already difficult
for a model to answer a question based on such a large context. To make it even more challenging, LV-Eval created two variants of the original datasets which we show in Figure \ref{fig:lveval}. In the \textbf{Confusing Facts Insertion (CFI)} variant, GPT-4 is prompted to generate sentences that are similar to the given question and the answer. These sentences are resolved for inconsistencies by human annotators and are then randomly placed in the original context. Owing to their similarity with the question and the answer, the purpose of the newly added sentences is to mislead the model. An example of CFI is illustrated in \ref{fig:lveval} where the original sentence refers to actor \textit{Tom Holland} and the Marvel movie \textit{No Way Home}. To confound the model, the newly generated sentence talks about a different actor, \textit{Tom Hiddleston} and his Marvel movie, \textit{Ragnarok}.
The \textbf{Keyword and Phrase Replacement (KPR)} variant is generated by selecting certain keywords or phrases and replacing them with other keyword or phrase throughout the context. This is done to ensure that the model is not reliant on its memorized prior knowledge while answering the given question. In Figure \ref{fig:lveval}, the KPR sentence is formed by replacing the actor \textit{Tom Holland} in the original sentence with the footballer \textit{Wayne Rooney}.

\begin{table*}[]
\centering
\begin{adjustbox}{width=0.85\textwidth}
\begin{tabular}{l|lllll|l}
\hline
{Method} &
  {16k} &
  {32k} &
  {64k} &
  {128k} &
  {256k} & 
  {mean}\\ 
  \hline
Mistral-7b-instruct-v0.2              &  75.6  & 56.3 & 40.6 & 32.5 & 21.9 & 45.4 \\ 
Phi-3-small-128k-instruct             & 65.6  & 65.6 & 64.4 & 46.2 & 30.0 & 54.4\\ 
\hline
Dragon + Mistral-7b-instruct-v0.2      & \textbf{78.1}  & 76.9 & 76.9 & 78.1 & 76.9 & 77.4\\ 
Dragon + Phi-3-small-128k-instruct     & 65.6  & 63.1 & 61.8 & 63.7 & 63.7 & 63.6\\
Dragon + Phi-3-small-128k-instruct     & 65.6  & 63.1 & 61.8 & 63.7 & 63.7 & 63.6\\
\hline
EpMAN (Uniform train - Best test)                        & 69.4  & 68.8  & 67.5  & 69.4  & 66.9  & 68.4  \\
EpMAN (Uniform train - Uniform test)                     & 71.3 & 71.9 & 70.6 & 72.5 & 72.5 & 71.8 \\
EpMAN (Uniform train - BroadAttn test)                   & \textbf{78.1} & \textbf{79.4} & \textbf{77.5} & \textbf{78.8} & \textbf{79.4} & \textbf{78.6} \\
EpMAN (Noisy train - Best test)                        & 70.6  & 72.5  & 70.6  & 70.0  & 70.6  & 70.9  \\
EpMAN (Noisy train - Uniform test)                      & 72.5 & 73.1 & 73.1 & 70.6 & 71.3 & 72.1 \\
EpMAN (Noisy train - BroadAttn test)     &  {75.6} & {77.5} & {76.3} & {75.0} & {75.0} & {75.9} \\ 
\hline
\end{tabular}
\end{adjustbox}
\caption{Performance of on loogle-SD-mixup using LLM-as-judge.}
\label{LOOGLE}
\end{table*}

\subsection{Baselines}
We choose two kinds of baselines for comparison with \textbf{EpMAN} described below:

\textbf{Baseline models}:
We first compare with instruction-tuned LLM decoders to investigate if they can generalize to longer context. While considering \texttt{Mistral-7b-instruct-v0.2}, following \cite{yuan2024lv}, we use half of the context from the top and half from the bottom in case the context size exceeds the model train context length. We also consider Phi-3-small-128k-instruct as a baseline model that was specifically trained with longer context inputs. 

\textbf{RAG}: We also compare with RAG systems to specifically evaluate if our \textbf{EpMAN} style training yields benefits over Retrieval Augmented Generation. We used state-of-the-art retrievers for instance Dragon~\cite{lin2023train, liu2024chatqa} with the above baseline decoder models.

\subsection{Results}
\subsubsection{Simple Recall Performance}

\textbf{NITH}: Table~\ref{NITH} shows the recall of \textbf{EpMAN} with other baselines for sentence completion NITH task. This is a simple recall task and \textbf{EpMAN} shows near perfect recall score on both the Paul Graham and PG-19 haystack cases showing that our decoder coupling with episodic attention can successfully complete the needle sentence when presented with partial information. The large context models, although trained at higher context length, can hallucinate and does not result in high recall. Using RAG with baseline decoders with Dragon retriever improves performance at higher context length although the recall is not perfect. Since this is a simple task, \textbf{EpMAN} with uniform and noisy training shows similar performance. Additionally, NarrowAttn and Exact methods show similar performance since information diffusion does not happen in this simple task.

\subsubsection{LV-Eval (CFI + KPR) performance}

\textbf{FactRecallEn}: From Table~\ref{FACTRECALL}, we observe that for the baseline models, the \texttt{Phi-3-small-128k-instruct} gets good performance on shorter context until 64k context size, however does not perform well for higher context lengths. Adding Dragon with these models improves long context performance, however \textbf{EpMAN} shows the best overall performance. It is important to note that \textit{Exact} inference does not perform well for this dataset because the Dragon encoder does not always extract the relevant context as the top entry, due to presence of CFI and KPR as described in Section~\ref{sec:eval_data}. Therefore, the relevant entry in the episode gets a low episodic attention score. Noisy training introduces robustness in the training, 
hence yields superior performance on longer context, even compared to uniform training due to the denoising objective.  \textit{BroadAttn} during inference provides additional performance boost at all context lengths.

\textbf{MultifieldQA}: As we move from recall tasks to complex QA tasks, it becomes more evident that \textit{BroadAttn} and noisy training improves the robustness of the decoder. Table~\ref{MFQA} shows that noisy-trained \textbf{EpMAN} with \textit{BroadAttn} get the best LLM-as-Judge score on this dataset compared to the other combinations. Similar to FactRecall-en, the presence of CFI and KPR, makes it difficult for the retriever to assign correct episodic weights in this challenging benchmark. Baseline models struggle to get competitive score; however, adding a retriever  to Mistral-7b instruct shows promising performance. However, since this decoder in simple RAG setup is not trained to expect noise in retrieval, noisy-trained \textbf{EpMAN} out-performs the RAG baseline.

\vspace{0.1cm}

\textbf{LoogleQA}: Table~\ref{LOOGLE} shows the performance on LoogleQA task. 
We observe Dragon + Mistral decoder gives the best performance among the baselines, while the non-RAG systems does not show competitive performance. For \textbf{EpMAN}, uniform training with \textit{BroadAttn} gives the best result outperforming the best baseline model. For Loogle~\cite{li2023loogle} we hypothesize that since some of the data is curated from Wikipedia, it might be in-domain compared to our synthetic training data, which is also sourced from Wikipedia. Additionally, the data for training the retriever is also derived from Wikipedia~\cite{lin2023train}. Therefore, the retriever might not add noise in this case, and consequently the denoising objective might not be necessary for robust response generation.

\section{Conclusions}
We present \textbf{EpMAN} - a novel method to generalize to long context using episodic attention in language models. Our method uses a two-level attention mechanism by first estimating the relative relevance of entries within a context and then re-weighting the self-attention scores for the entries by the corresponding episode-level relevance score. Our architectural improvements - differentiating attention and training with denoising objective - show robust performance on complex recall and question answering tasks in the presence of distracting information and replaced keywords. Additionally, our attention scope expansion during inferences also proves to be beneficial in such challenging settings especially for QA tasks.

\section{Limitations}
While our method proposes techniques to improve attention over relevant chunk, we store the full KV cache for this work, which would take large CPU/GPU memory for large document processing and might require more processing time for CPU to GPU transfers (when we store the KV cache in CPU). Furthermore, using a large top K value for episodic attention would also requires more memory for training, especially large models. Additionally, another limitation is that the benefits of uniform/noisy training and exact/narrow/broad attention depends on the nature and complexity of the task. We plan to introduce methods like KV cache compression and pruning to make our approach more scalable and efficient in future works.

\bibliography{anthology,custom}

\newpage

\appendix

\begin{table*}[t]
\centering
\begin{adjustbox}{width=0.8\textwidth}
\begin{tabular}{l|ccccc|c}
\hline
Method                             & 16k  & 32k  & 64k  & 128k & 256k & Mean \\ \hline
Mistral-7b-instruct-v0.2           & 65.3 & 72.5 & 41.0 & 22.5 & 11.5 & 42.6 \\
Phi-3-small-128k-instruct          & {82.0} & {80.5} & {81.0} & 63.0 & 34.5 & 68.2 \\ \hline
Dragon + Mistral-7b-instruct-v0.2  & 74.2 & 71.8 & 66.0 & {77.2} & 69.0 & 71.7 \\
Dragon + Phi-3-small-128k-instruct & 71.8 & 70.5 & 68.0 & 76.0 & 68.5 & 71.0 \\ \hline
EpMAN (Noisy train - BroadAttn test)   & 81.8 & 75.2 & 76.0 & 75.2 & \textbf{80.2} & {77.7} \\
EpMAN (+ trainable read/write)  &   \textbf{83.0}  &   \textbf{88.5}   &  \textbf{89.0}  &  \textbf{89.0}  &  78.0  & \textbf{85.5} \\ \hline

\end{tabular}
\end{adjustbox}
\caption{Performance of various models on Factrecall-en using recall metric. EpMAN with noisy training with BroadAttn shows overall best performance.}
\label{FACTRECALL_trainable}

\end{table*}

\begin{table*}
\centering
\begin{tabular}{|l|r|r|r|r|r|r|}
\hline
\textbf{Method} & \textbf{Total} & \textbf{C0} & \textbf{C1} & \textbf{C2} & \textbf{C3} & \textbf{C4} \\
\hline
Topic Sampled & 171576 & 34200 & 34262 & 34280 & 34362 & 34472 \\
Wikipedia & 89253 & 17812 & 17636 & 18257 & 17765 & 17783 \\
\hline
\end{tabular}
\caption{Details of synthetic data used for training EpMAN}
\label{synth_data_details}
\end{table*}

\begin{table*}[t]
\centering
\begin{adjustbox}{width=0.8\textwidth}
\begin{tabular}{l|c|ccccc|c}
\hline
Method              & Top K               & 16k  & 32k  & 64k  & 128k & 256k & Mean \\ \hline
EpMAN (Uniform train - BroadAttn test) & 2 & \textbf{88.2} & 82.8 & 78.2 & 79.5 & \textbf{82.5} & 82.2 \\
EpMAN (Uniform train - BroadAttn test) & 3 & 82.0 & 73.0 & 71.5 & 70.0 & 79.0 & 75.1 \\
EpMAN (Noisy train - BroadAttn test) & 2 & 87.0 & \textbf{86.0} & \textbf{80.5} & \textbf{80.2} & 82.2 & \textbf{83.2} \\
EpMAN (Noisy train - BroadAttn test) & 3 & 81.8 & 75.2 & 76.0 & 75.2 & 80.2 & 77.7 \\ \hline

\end{tabular}
\end{adjustbox}
\caption{Performance of various top K values on factrecall-en dataset using recall metric. Top K value of 2 with noisy training gives the best score for this dataset}
\label{topk_factrecall}
\end{table*}

\section{Effect of trainable memory operations}
In the previous experiments, we used the setting of a fixed retriever using Dragon~\cite{lin2023train} for fair comparison with the baseline methods. However, as we described in Section~\ref{sec:epman_details}, our memory operations can also be trained using the loss described in Equation 2. We trained \textbf{EpMAN} in a two phase manner where in phase 1 we train the memory operations (read and write are single layer MLPs) and a BGE~\cite{chen2024bge} retriever. We divide our training data into different samples phase 1 and 2 respectively. Once we train the first phase, we use the train memory operation to obtain the episodic attention on the chunks. In phase 2, only the decoder parameters are updated.

Table~\ref{FACTRECALL_trainable} shows the performance of the \textbf{EpMAN} with trained memory operations. Since the phase 1 training improves the retriever performance for obtaining the relevant chunk from the large context, the trained decoder can generate accurate responses leading to improved performance compared to fixed retriever setting. However, we do not report this in the main paper because the RAG baselines should also use such an improved retriever for fair comparison. It should be noted that although we are using a trained retriever + memory operations in this setting, the data we are evaluating is out-of-distribution. To improve the performance on a large variety of OOD data, we can train the retriever using the retriever dataset in addition to our dataset using contrastive learning.

\section{Effect of top K for BroadAttn}
In this experiment, we look at the effectiveness of using BroadAttn for various values of top K. We used a default value for top K as 5 for the experiments in the main paper (except for factrecall-en where we use top K value of 3). 
Table~\ref{topk_factrecall} show the performance on factrecall-en for various top K values. We find that a top K value of 2 performs better for BroadAttn. We hypothesize that since BroadAttn includes the relevant context neighbors it might add some distractor chunks that might confuse the decoder. Therefore, having a lower top K would reduce the number of such distractor chunks leading to better performance. However, it should be noted that for complicated QA datasets, where the retriever might not be able to pick the relevant context in a smaller top K setting, it might lead to worse performance. Therefore, this analysis might not be general and might vary based on the complexity of the dataset.

\section{LLM-as-Judge Prompt}
We evaluate the performance of various models on MultiFieldQA and LoogleQA using LLM-as-Judge~\cite{zheng2023judging}. The existing metrics that was used in LVEval~\cite{lveval} did not account for variation in length in the answers compared to the gold. Also, those metrics did not consider rephrases in the answers. Therefore, we found, although the answers were correct (albeit full sentence responses), the F1 score metric did not reflect that. Therefore, we used \textsc{mistralai/Mixtral-8x22B-Instruct-v0.1} to compare the generated responses with the gold response. Figure~\ref{llm-as-judge} shows the prompt we used for this purpose.

\label{sec:appendix}

\begin{figure}[h]
\centering
\small
\begin{center}
\begin{verbatim}
  <s>[INST]
  You are tasked as an expert language model 
  judge to analyze two answers 
  from different sources.
  Your objective is to determine how similar
  they are.
  Provide a score based on their correspondence:

  Score of 0 (Zero): Assign this score if the 
  answers discuss different things 
  and are unrelated.

  Score of 1 (One): Assign this score if the 
  answers are similar or have a common 
  theme or topic in common.

  Issue your final score as:

  FINAL SCORE: 0 for mismatched passages.
  FINAL SCORE: 1 for matched passages.

  Here is the first answer: <<generated answer>>.
  And here is the second answer: <<gold answer>>.
  Now go ahead and provide your final final score,
  accurately reflecting similarity.
  Make sure to use the format "FINAL SCORE: 
  [your score]" as your only output.
  Skip the preamble and provide only the final score.
  [/INST]
\end{verbatim}
\end{center}
\caption{LLM-as-Judge prompt that was used to measure the performance of the MultiFieldQA and LoogleQA}
\label{llm-as-judge}
\end{figure}

\section{Details about the synthetic data}
In addition to the description in Sec~3.3.2 in the paper, we provide additional details of our synthetic data. For each type of synthetic dataset, we add hard-negatives in the training set. We show the total number of training samples and the number of samples with $m$ hard-negative given by C$m$ column in Table~\ref{synth_data_details}. For checking the consistency of the answer with the context, we used factuality model (\textsc{nightdessert/WeCheck}) to ensure that the generated passage, questions and corresponding answers are factually consistent. In addition, we also manually checked random samples from the synthetic data to ensure that generated samples are of sufficient high quality. For the synthetic data based on wikipedia, since the passages are sampled from wikipedia they are human generated and not synthetically generated. In this case, only the question and answer is generated using the teacher model. Similar to previous case, we also perform manual consistency check of random samples from the data to ensure they are of high quality.

We show an example of a topic sampled generated data with a hard negative below:

\begin{itemize}
\item \textbf{Synthetic Question:} What is the estimated age of the deposits in which the fossils of Cystoides estonicus were found, according to the passage?
\item \textbf{Synthetic Context:} Recent discoveries in the field of paleontology have shed new light on the ancient crustacean group Cystoidea, with the unearthing of exceptionally preserved fossils in the Upper Ordovician deposits of Estonia. The newly described species, Cystoides estonicus, boasts an extraordinary level of ornamentation, featuring intricate patterns of ridges and tubercles on its calcite shell. Measuring up to 10 centimeters in diameter, these ancient echinoderms are believed to have played a crucial role in the Ordovician ecosystem, serving as both predators and prey for other marine organisms. The remarkable preservation of soft tissues in these fossils has also provided valuable insights into the anatomy and possible feeding mechanisms of these enigmatic creatures. Further study of Cystoides estonicus is expected to significantly expand our understanding of the evolution and diversification of cystoids during the Paleozoic Era.
\item \textbf{Answer:} The age of the deposits in which the fossils of Cystoides estonicus were found is from the Upper Ordovician period.
\item \textbf{Hard negative:} In the course of time, however, a shift can be observed in the temporal significance of these terms, from post- Eocene to post-Early Miocene to post-middle Pleistocene. The region is seismically active and is generally ascribed to the re-establishment of an equilibrium after the latest (mid-Pleistocene) deformation phase. Some authors believe that the subduction process is still ongoing, which is a matter of debate. History. Calabria has one of the oldest records of human presence in Italy, which date back to around 700,000 BC when a type of \" Homo erectus \" evolved leaving traces around coastal areas. During the Paleolithic period Stone Age humans created the \" Bos Primigenius \", a figure of a bull on a cliff which dates back around 12,000 years in the Romito Cave in the town of Papasidero. When the Neolithic period came the first villages were founded around 3,500 BC. Antiquity.
\end{itemize}

\section{Episodic attention and training}

We provide additional details about $a_{mem}$ based transformation of self attention (described in Section 3.2 and 3.4). Suppose, we have a document/context with $N$ tokens -- the corresponding KV dimensions are $N X 32 X 128 X 8$ (for each K and V matrix) for Mistral model since it has the head dimension of 128 and 8 multi-heads and 32 layers. We do not mention the batch dimension here and assume B=1 for simplified explanation. Assuming that each chunk has a length of 256 tokens, there are $N_{ep}$ number of chunks where $N_{ep} = N/256$. We obtain chunk-level episodic weights (lets call it $a_{mem}^{chunk}$ vector) of size $N_{ep}$. Since the softmax score, given by $x_{softmax} = \frac{softmax(qK^T)}{\sqrt{d_z}}$, is of size $N$ (it attends to the whole context), we broadcast the $N_{ep}$ sized $a_{mem}^{chunk}$ vector to $N$ sized vector $a_{mem}$ vector, such $a_{mem}[k*256: (k+1)*256] = a_{mem}^{chunk}[k]$. Subsequently, $a_{mem}$ is used to reweight the $x_{softmax}$ which is then multiplied with the value vector to get the final context vector.

In our implementation, we use random sampling between $[1, 1-\beta K]$, where $\beta$ signifies the slope of episodic attention attenuation. This setup enforces the setting that as the number of top K chunks increases, some of the chunks should get high weights while other chunks would get lower weights. This would emphasize diversity of $a_{mem}$ weights mimicking retrieval weights during inference on OOD dataset (explained in Sec 3.4). By using random sampling, we ensure that the model does not always assign high weights to the top chunks for in-distribution training; otherwise training the decoder with the original retrieval weights would lead to overfitting and poor generalization to OOD datasets. For our experiments, we use $\beta=0.2$ and therefore we sample from $[1.0, 0.9]$ since we use top K = 5 during training. This setup enables the decoder to attend to chunks that do not share the highest similarity with respect to the question during inference, but still are of high relevance. As a result, the OOD generalizability of the EpMan attention is enhanced, as the decoder is not trained to only attend to the chunks of highest similarity seen during training.

\end{document}